\documentclass{article}

\usepackage{arxiv}

\usepackage[utf8]{inputenc} 
\usepackage[T1]{fontenc}    
\usepackage{hyperref}       
\usepackage{url}            
\usepackage{booktabs}       
\usepackage{amsfonts}       
\usepackage{nicefrac}       
\usepackage{microtype}      
\usepackage{lipsum}
\usepackage{graphicx}

\usepackage{amsmath,amssymb,amsfonts}
\usepackage{algorithmic}
\usepackage{textcomp}
\usepackage{hhline}
\usepackage{booktabs}
\usepackage{multirow}
\usepackage{color,colortbl}
\usepackage{subfigure}

\usepackage{geometry}
\usepackage{pdflscape}
\usepackage{enumerate}

\title{Handling Missing Annotations in Supervised Learning Data}

\author{
  Alaa E. Abdel-Hakim \\
  Department of Computer Science\\
  Umm Al-Qura University\\
  Jamoum, Saudi Arabia, 25371 \\
  Electrical Engineering Department \\
  Assiut University \\
  Assiut, Egypt, 71516 \\
  \texttt{alaa.aly@eng.au.edu.eg} \\
   \And
 Wael Deabes \\
  Department of Computer Science\\
  Umm Al-Qura University\\
  Jamoum, Saudi Arabia, 25371 \\
  Department of Computers and Systems Engineering\\
  Mansoura University\\
  Mansoura, Egypt, 35516 \\
  \texttt{wadeabes@uqu.edu.sa} \\
}

\begin{document}
\maketitle
\begin{abstract}
Data annotation is an essential stage in supervised learning. However, the annotation process is exhaustive and time consuming, specially for large datasets. Activities of Daily Living (ADL) recognition is an example of systems that exploit very large raw sensor data readings. In such systems, sensor readings are collected from activity-monitoring sensors in a 24/7 manner. The size of the generated dataset is so huge that it is almost impossible for a human annotator to give a certain label to every single instance in the dataset. This results in annotation gaps in the input data to the adopting supervised learning system. The performance of the recognition system is negatively affected by these gaps. In this work, we propose and investigate three different paradigms to handle these gaps. In the first paradigm, the gaps are taken out by dropping all unlabeled readings. A single \textit{``Unknown''} or \textit{``Do-Nothing''} label is given to the unlabeled readings within the operation of the second paradigm. The last paradigm handles these gaps by giving every one of them a unique label identifying the encapsulating deterministic labels. Also, we propose a semantic preprocessing method of annotation gaps by constructing a hybrid combination of some of these paradigms for further performance improvement. The performance of the proposed three paradigms and their hybrid combination is evaluated using an ADL benchmark dataset containing more than $2.5\times 10^6$ sensor readings that had been collected over more than nine months. The evaluation results emphasize the performance contrast under the operation of each paradigm and support a specific gap handling approach for better performance.
\end{abstract}


\section{Introduction}
Learning systems take decisions based on gathering experiences embedded in existing data \cite{Li2015a}. However, an amount of labeled data intensely affects the performance of such systems. Learning methods are classified according to the availability of the labeled training data to supervised and unsupervised learning \cite{Weber2017}. Acquiring complete labeled data for machine learning is regularly very challenging and prohibitive, therefore involving unlabeled data in the supervised learning process is vastly promising, since it expands the accuracy of those learning methods \cite{Trabelsi2013}. Moreover, a mutual hypothesis in supervised learning is that the collected labeled data represents a normal distribution of process characteristics. However, labeled datasets are frequently small due to the expensive labeling cost and/or sampling bias, which mislead the classification process and produce a non-generalized classifier~\cite{Toda2014}.

Recently, hybrid learning methods that can learn from labeled and unlabeled data gained much interest hoping to construct better performing classifiers~\cite{Yoon2011}. These methods can be categorized into three major approaches varying from fully supervised to unsupervised learning \cite{Tan2017}: (1) Building an initial classifier based on a set of labeled data then use it for labeling the unlabeled data. Hereafter, a new classifier is built based on both the earlier and the new labeled data. Techniques such as a Self-Organizing Map (SOM), neural network \cite{Dara2002} and \lq co-training\rq~using Na\"{i}ve Bayes classifiers or Expectation Maximization (EM) are usually applied in this first approach \cite{Lee2014}. (2) The second approach usually generates a data model using all available data by applying either a data density estimator or clustering procedure. Then, the labels are consequently applied to label entire clusters of data or estimate class densities. These class densities comprise labeling of the unlabeled data based on their relative position in the data space with respect to the originally labeled data \cite{Pentina2016}. One of the known techniques in this category is the probabilistic framework, which applies a mix of Gaussians or Parzen windows for learning \cite{Chawla2005a}. (3) Both labeled and unlabeled data are processed together for creating semi or partially supervised classifier. This category is between the first and the second categories, since clustering depends on a proper similarity measure and is directed by labeled data. One of the techniques that belongs to this category is a General Fuzzy Min-Max (GFMM) neural network which iteratively processes both labeled and unlabeled data for adjusting hyper box fuzzy clusters \cite{Gabrys2000}. Fairly~diverse methods belong to this group are presented in \cite{Cohn2003}. Comparison between the classifiers based on only limited labeled data and those discussed above which use further unlabeled data shows the feasibility and the advances of the former techniques.

One of the well-known methods for conducting unlabeled data is the Expectation Maximization (EM) algorithm. EM handles the unlabeled data as missing data and assuming the generative model such as a mixture of Gaussians for iteratively estimating the model parameters and assigning soft labels to unlabeled examples. Applications such as text classification~\cite{Kowsari2017}, image retrieval~\cite{Vrigkas2015} bioinformatics~\cite{Szilagyi2010}, and natural language processing~\cite{Goutte2002} widely use EM. Co-training approach \cite{Goldman2000} is another common method for integrating unlabeled data, if the data can be classified into two different sets of attributes. Instead, the transduction approach maximizes the classification margins on both labeled and unlabeled data to generate labels to a set of unlabeled data. Nonetheless, it is commonly observed that these methods could reduce performance due to abused model assumptions or sticking in local maxima~\cite{Seeger2001}. 

However, many empirical studies show that using unlabeled data does not always improve the performance of some classifiers \cite{Li2015a}. Thus, such facts definitely burden the deployment of semi-supervised learning in high-reliability real applications, which require more accurate performance compared with the existing supervised techniques. Therefore, it is crucial to build a safe semi-supervised classifier using unlabeled data without reducing the performance significantly~\cite{Tan2017}. The word safe means that the overall performance is not statistically considered worse than those using only labeled data. Many trials must be conducted, since the performance might be worse by exploiting more labeled data. The reason for the cruel effect of the unlabeled data is discussed in~\cite{Cozman2002}. They~inferred that the incorrect model assumptions are the main cause of the performance worsening since the learning process will be deceived by the wrong model assumptions. However, it is hard to create accurate model assumptions without applicable domain knowledge. Instead, methods that disagree with this belief usually use various learners to provide pseudo-labels of unlabeled data to enrich the whole data set. However, wrong pseudo-labels may confuse the learning process, but~usually, data editing techniques are used to overcome this problem particularly on dense data, since these techniques trust on the data adjacent information~\cite{Wu2006}.

This paper attacks the problem of studying and recognizing human activities of daily living (ADL), which is an important research issue in building a universal and smart environment. Consequently, it is crucial to build semantically rich models permitting knowledge in real time of which activity is performed by an occupant living in a smart home, from semantically poor sequences of observed binary events. Moreover, the main objective of this work is to find the best way to handle annotation gaps in the raw input human activity data.  Therefore, we investigate the problem of unlabeled data from different perspectives using proposed annotation gaps handling paradigms. We look at the annotation gaps as unknown data that need to be used in a way that assists performance improvement. First, in paradigm~\#1, these gaps are treated as a source of confusion and removed. In the second paradigm, all gaps are classified as a unique \textit{``Unknown''} class. The third paradigm characterizes the gaps by their encapsulating classes. This paradigm is inspired by observed input data pattern repetition depending on preceding and consequent activities. Finally, the first and the third paradigms are combined in a preprocessing stage to exploit some prior knowledge about the physical nature of the ground truth activities for further performance improvement. Experimentally based on a real smart home dataset, the proposed three paradigms are applied to evaluate the effect of each of them on different activity recognition metrics.

The rest of this paper is organized as follows: in Section~\ref{sec:HMM}, an overview of the used classifier, Hidden Markov Model, is presented. In Section~\ref{sec:gaps}, the problem of annotation gaps with the reasons behind are discussed. The proposed annotation gap handling paradigms are detailed in Section~\ref{sec:gap_handling}. The experimental setup is explained, and the evaluation results are discussed in Sections~\ref{sec:setup} and~\ref{sec:discussion}, respectively. Finally, the paper is concluded in Section~\ref{sec:conclusion}.


\section{Hidden Markov Modeling of Activity Recognition}
\label{sec:HMM}
Currently, many mathematical models for activity recognition system have been developed, such as Hidden Markov Model (HMM) \cite{Liisberg2016}, Bayesian Network \cite{Oliver2005}, Fuzzy Systems \cite{Banerjee2014} and Neural Network \cite{Bourobou2015}. Each model has its features and benefits. However, none of them can be assured to be entirely accurate, truthful, and totally free of error. However, HMM represents the most activity recognition model that is easy to develop and deploy. 

\textls[-15]{HMM is one of the simplest forms of the Dynamic Bayesian Networks (DBNs), where it has one discrete hidden and observed nodes per segment. When dealing with spatio-temporal information, the~accuracy of the HMMs is superior. Therefore, most of the activity recognition literature is based on HMMs. The main components of the HMM are represented by 5-tuple $\lambda=(S, V, A, B,\pi)$ as follows \cite{Rabiner1989}:}

\begin{enumerate}
\item $S={s_1,s_2 \dots s_N}$  is the set of the hidden states, $N$ is the number of the states. The active state at a time instant, $t$, is $q_t$. In the activity recognition model considered herein, the states represent the conducted activity in this specific instant, $t$;

\item $V={v_1,v_2 \dots ,v_M}$ is the set of observable events and $M$ is the number of observation events. The~observable events in our case are the sensor readings;

\item \textls[-15]{$A= {a_{ij}}$ is the state transition probability distribution, where $a_{ij}=P[q_{t+1}=s_j | q_t=s_i],   1\leq i,j \leq N$;} 

\item     $B= {b_j (k)}$ is the observation event probability distribution for state $s_j$ where $b_j (k)=P[v_k \,\, at \,\,  t | q_t=s_j ],     1 \leq j \leq N,1 \leq k \leq M$;

\item     \textls[-20]{$\pi=\{\pi_1,\pi_2 \dots \pi_N\}$ is the set of initial state probability distribution where $\pi_i=P[q_1=s_i],   1 \leq i \leq N$.}
\end{enumerate}

\begin{figure}
  \centering
 \includegraphics[width=.85\textwidth]{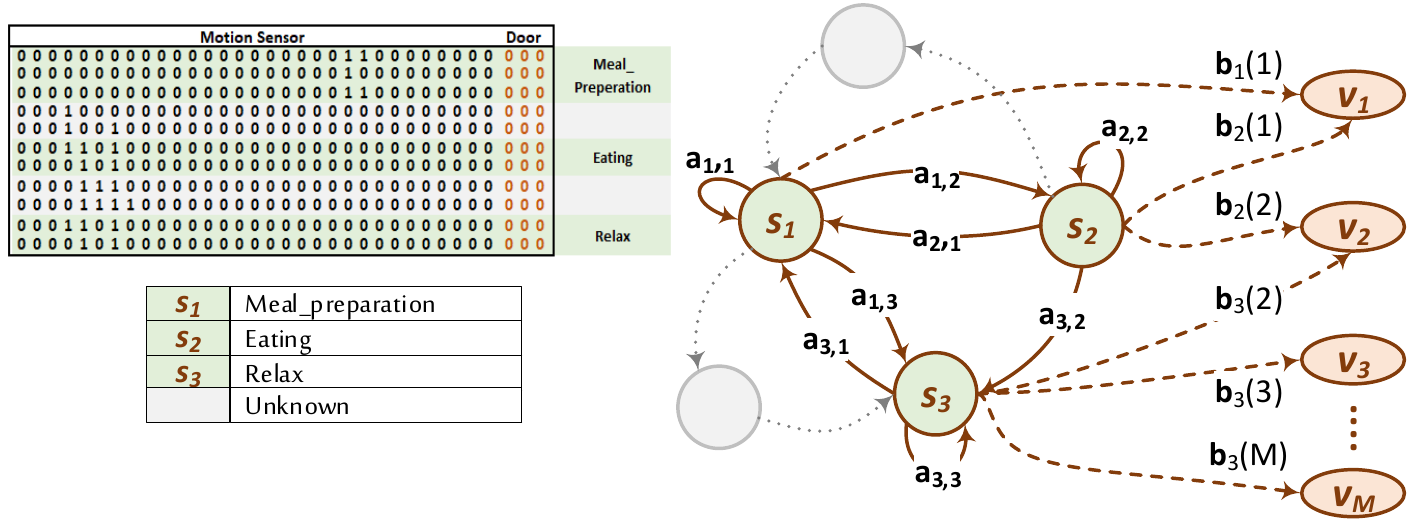}
\caption[\textwidth]{Graphical representation of the used HMM for activity recognition. Annotated events are represented by solid-circle states. Annotation gaps are shown in dimmed circles.}
\label{fig:HMM}
\end{figure}

Figure \ref{fig:HMM} shows the structure of an HMM. For illustration purposes, only three activities along with two gaps are shown. The annotated activity transitions are represented by solid arcs, $a_{ij}$, while~the activity gaps are dimmed out.

For modeling the activity recognition problem by HMM, any activity is represented by a sequence of hidden states. At any time $t$, the user is assumed to be at one of these states and each state releases events with certain observation event probabilities $b_j(k)$. In the following time slot, $t+1$, the system goes to another state, resulting in another activity, according to the transition probabilities between the states~$a_{ij}$. During the training phase, an annotated dataset is used to calculate the transition and observation event probabilities, transition, and emission models respectively, by solving decoding problem, i.e., estimating the most expected state sequence that achieves the events \cite{Aggarwal2011}. In the classification stage, the estimated transmission and emission models are used to estimate hidden states, which are the sought activities.


\section{Annotation Gaps}
\label{sec:gaps}

At the core of activity recognition systems, multi-modal data, which is acquired by various sensors, represents the main ingredient for recognition algorithms. Typically, sensors are installed around target premises or house to ``continuously'' monitor the household activities. These sensors are physically installed in locations, which guarantee a collection of data assisting activity recognition algorithms. To provide such assistance, collected data must achieve a tradeoff between distinction and compactness. In other words, the collected data should be as compact as possible to get rid of any reading redundancy~\cite{hakim2017impact}. Yet, the collected readings should contain data that can be used to clearly distinguish between different activities. Types of sensors, readings resolution, installation positions, and operation environments usually contribute to the achievement of such tradeoff.

Used sensors may be visual, such as cameras of different modalities, or non-visual such as temperature and motion sensors. The scope of this paper is handling the data acquired by non-visual sensors. Typically, non-visual sensors provide digital and/or analog readings 24/7. The richness of such data could be an advantage in many cases and a source of challenge in other cases~\cite{hakim2017impact}. One of the challenges created by the big data size shows up in feeding such data into supervised learning-based frameworks. In particular, getting a ground truth label for every single instance in the acquired data is an extremely exhaustive process, if not impossible. This is because of the dependence on common manual labeling procedures.

A human annotator finds difficulties in giving a crisp activity label to each single reading instance due to several reasons such as the following:

\begin{enumerate}
\item There is a usual delay between acquisition and annotation. This results in missing specific labels of some activities.
\item By nature, many activities lack clear in-between boundaries determining their beginnings and their ends.
\item A household himself/herself is often the annotator. He/She usually gives labels to activities on fuzzy basis depending on his/her memory of the starts and the ends of the activities.  Supported by the aforementioned annotation delay, this leads to consequent annotation inaccuracy. 
\end{enumerate}

All these factors lead the annotator to leave some unlabeled gaps surrounded by well-determined activity readings. For training processes, these gaps represent a source of modeling inaccuracy. Since~it is extremely difficult to avoid the creation of such label gaps in the training data, we investigate various approaches to handle them during the training process. In the next section, we propose three different paradigms to handle these gaps. 


\section{Gap Handling Paradigms}\label{sec:gap_handling}
In this section, we explain the proposed paradigms for dealing with existing annotation gaps in the acquired sensor data.

\subsection{Gap Removal}
\begin{figure}
  \centering
 \includegraphics[width=.85\textwidth]{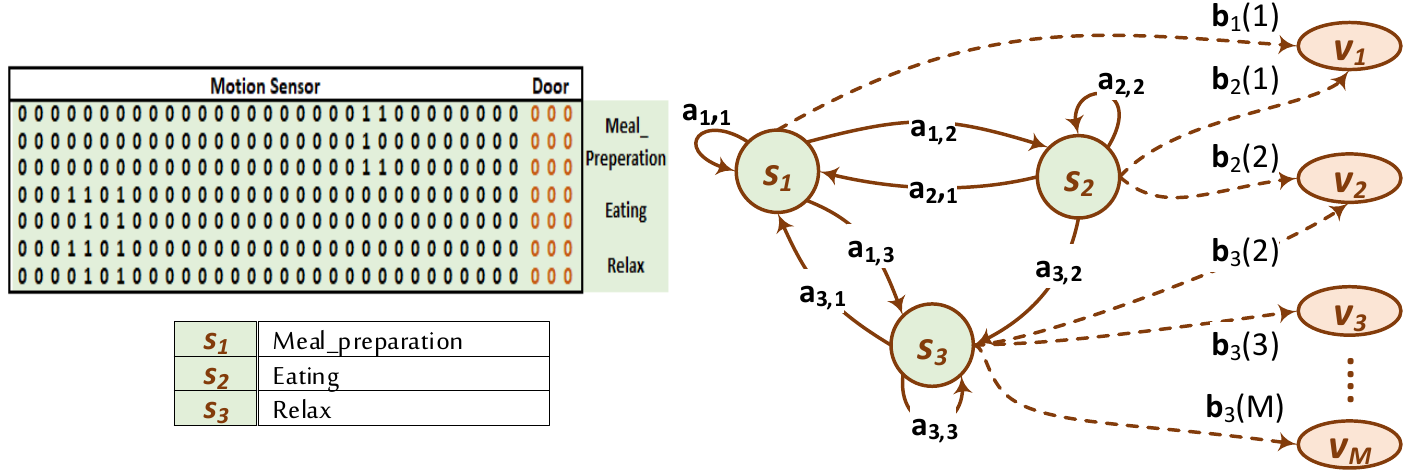}
\caption[]{Paradigm~\#1: Annotation gaps are removed completely from the training data.}
\label{fig:All2}
\end{figure}

In this paradigm, all unannotated data are dropped. In this case, the transition model will look at the input activities as if they are connected sequences. The input data will turn out to be a compressed version, which truncates the unannotated readings and consider the human subject in an idle state that is not taken into consideration for model building. Figure~\ref{fig:All2} shows an example of the input data after the application of this approach. In this case, the HMM model parameters are modified as follows:
\begin{equation}
\label{eq:trans_cond}
    a_{ij} = 0;\; \forall s_j = Null,\; 1\leq i\leq N
\end{equation}
\noindent where $s_j = Null$ indicates an annotation gap. Similarly, 
\begin{equation}
\label{eq:emission_cond}
    b_j(k) = 0;\; \forall s_j = Null,\; 1\leq k\leq M
\end{equation}

\textls[-20]{The transition and emission models are re-estimated after normalizing both the state transition and the observation event probability distributions according to the conditions of Equations~(\ref{eq:trans_cond}) and~(\ref{eq:emission_cond}), respectively.}

\subsection{A Unique Label for Gaps}
Unannotated data can be generated as a result of one of two cases: (1) undefined or idle user activities, (2) lack of ground truth provided by the human annotation. It is expected that removing these unlabeled data gaps, as illustrated in the previous approach, will work with the first case. However, the second case is common because of the difficulty of annotating all the collected sensor data. This~difficulty comes from the fact that most of the annotators are usually the residents themselves, who find the annotation of such huge-size sensor data a boring and a time-wasting process. In this case, dropping these unannotated data does not only block valuable information from the classification, but~it will also cause ``confusion'' and miss-training to the modeling process as well.

\begin{figure}
  \centering
 \includegraphics[width=.85\textwidth]{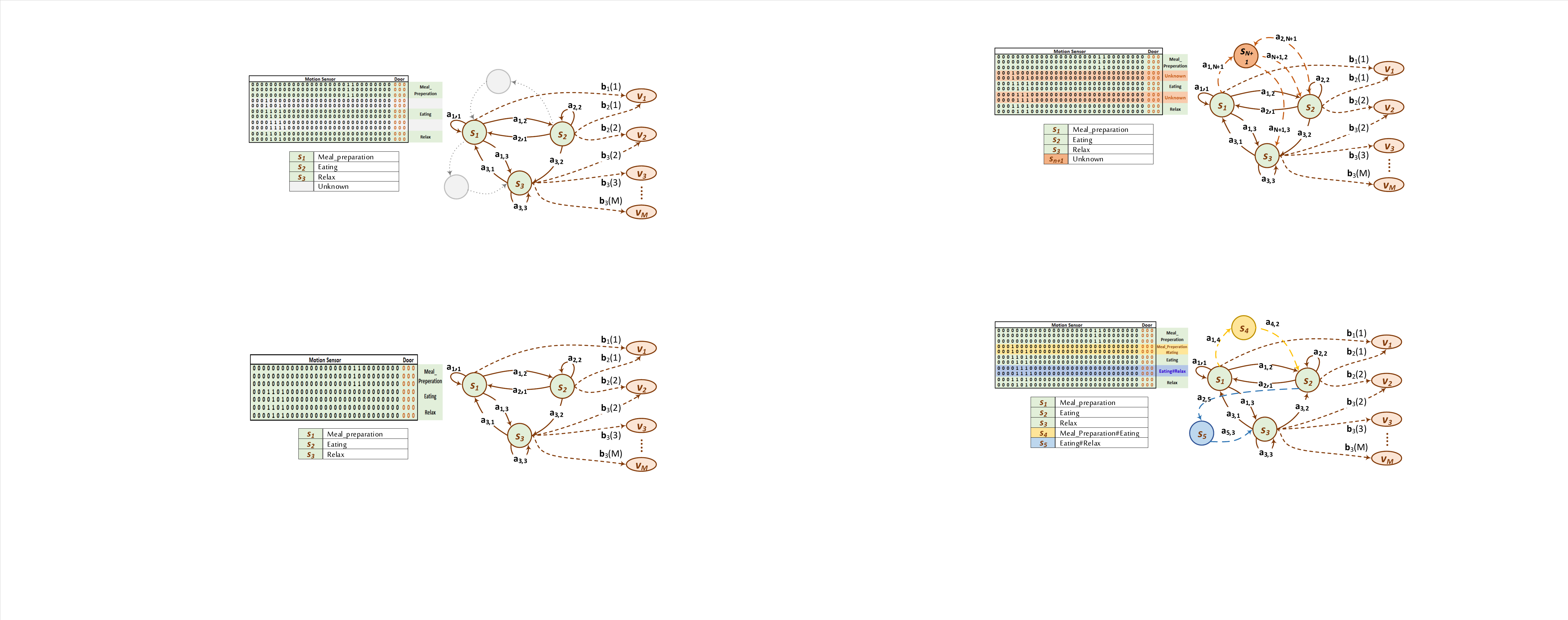}
\caption{Paradigm~\#2: Annotation gaps are assigned a unique \textit{``Unknown''} label under state $s_{N+1}$.}
\label{fig:All3}
\end{figure}

Therefore, to reflect the existence of some unknown activities between the annotated ones to the transition model, we give all these unannotated labels a single universal one, called ``\textit{unknown}''. This~``\textit{unknown}'' activity is plugged into the model exactly the same as any ordinary activity. \mbox{In this case}, the ground truth activity labels, $N$, are increased by one. In other words, the training process is designed for $N+1$ and $s_{N+1}$ is the\textit{ ``Do-Nothing''} label, as follows:
\begin{equation}
    S_{p2}= s_1,s_2,\dots,s_N,s_{N+1};\;s_{N+1}=``Unknown''
\end{equation}
\noindent \textls[-15]{where $S_{p2}$ is the state space of paradigm~\#2. Figure~\ref{fig:All3} illustrates inserting this new \textit{unknown} state into the HMM state space. It is considered in estimation of both transition and the emission models of the HMM}.

\subsection{Distinct Interactivity Labels}
A close look at the raw sensor readings gives an intuition of presence of some repeated pattern depending on the preceding and the posterior confirmed activities. These patterns can be considered to be identifying signatures of their encapsulating activities. For example, Table~\ref{tab:addlabel} shows exemplar readings captured from a real activity recognition test-bed which contains several 34 digital sensors. In the left half of the table, the preceding activity is \textit{``Sleeping''} , while the consequent activity is \textit{``Bed\_to\_Toilet''}. The right half shows similar sample readings for the same two activities but with reversed order. Existence of repeated patterns in sensor readings in both cases is clear. However, while the sensor readings in the first gap is oscillating between 0x0001 and 0x292D, readings of 0x0001, 0x2FA0, and 0x3459 repeatedly shows up in the second gap. This implies that not only the gap's encapsulating activities, but also their order, produce distinguishing signatures for the observable events within the gaps.

\begin{table}
  \centering
  \caption{Interactivity sensor readings in the annotation gaps.}

    \scalebox{.7}[.7]{\begin{tabular}{cccrccc}
\toprule    \rowcolor[rgb]{.557,.663,.859} \multicolumn{3}{c}{\textbf{Sleeping}} & \cellcolor[rgb]{1,1,1} & \multicolumn{3}{c}{\cellcolor[rgb]{.663,.816,.557}\textbf{Bed\_To\_Toilet}} \\
\midrule    \multicolumn{1}{c}{ \cellcolor[rgb]{.706,.776,.906} \textbf{Binary}} & \multicolumn{1}{c}{ \cellcolor[rgb]{.706,.776,.906}\textbf{Hex}} &     & \cellcolor[rgb]{1,1,1} & \multicolumn{1}{c}{\cellcolor[rgb]{.776,.878,.706}\textbf{Binary}} & \multicolumn{1}{c}{\cellcolor[rgb]{ .776,  .878,  .706}\textbf{Hex}} & \cellcolor[rgb]{.776,.878,.706} \\
\midrule  \multicolumn{1}{l}{\cellcolor[rgb]{1,.949,.8} 0000000000000000000000000000000001} & \multicolumn{1}{l}{\cellcolor[rgb]{ 1,  .949,  .8} 0x000000001} & \multirow{6}{*}{\cellcolor[rgb]{ 1,  1,  1}\textbf{Unknown}} & \cellcolor[rgb]{ 1,  1,  1} & \multicolumn{1}{l}{\cellcolor[rgb]{.988,.894,.839}0000000000000000000010111110100000} & \multicolumn{1}{l}{\cellcolor[rgb]{.988,.894,.839}0x000002FA0} & \multirow{2}{*}{\cellcolor[rgb]{ 1,  1,  1}\textbf{Unknown}} \\
  \multicolumn{1}{l}{\cellcolor[rgb]{1,.949,.8} 0000000000000000000010100100101101} & \multicolumn{1}{l}{\cellcolor[rgb]{1,.949,.8} 0x00000292D} &  & \cellcolor[rgb]{1,1,1} & \multicolumn{1}{l}{\cellcolor[rgb]{ .988,.894,.839}0000000000000000000000000000000001} & \multicolumn{1}{l}{\cellcolor[rgb]{.988,.894,.839}0x000000001}  & \\
 \multicolumn{1}{l}{\cellcolor[rgb]{1,.949,.8} 0000000000000000000000000000000001} & \multicolumn{1}{l}{\cellcolor[rgb]{1,.949,.8} 0x000000001} &  & \cellcolor[rgb]{1,1,1} & \multicolumn{3}{c}{\cellcolor[rgb]{ .886,.937,.855}\textbf{Sleeping}} \\
  \multicolumn{1}{l}{\cellcolor[rgb]{1,.949,.8} 0000000000000000000010100100101101} & \multicolumn{1}{l}{\cellcolor[rgb]{1,.949,.8} 0x00000292D} &  & \cellcolor[rgb]{1,1,1}&\multicolumn{3}{l}{\cellcolor[rgb]{1,1,1}}  \\
  \multicolumn{1}{l}{\cellcolor[rgb]{1,.949,.8} 0000000000000000000010100100101101} & \multicolumn{1}{l}{\cellcolor[rgb]{1,.949,.8} 0x00000292D} &     & \cellcolor[rgb]{1,1,1} & \multicolumn{3}{c}{\cellcolor[rgb]{ .663,.816,.557}\textbf{Bed\_To\_Toilet}} \\
    \multicolumn{1}{l}{\cellcolor[rgb]{1,.949,.8} 0000000000000000000000000000000001} & \multicolumn{1}{l}{\cellcolor[rgb]{1,.949,.8} 0x000000001} &     & \cellcolor[rgb]{1,1,1} & \multicolumn{1}{l}{\cellcolor[rgb]{ .988,.894,.839}0000000000000000000010111110100000} & \multicolumn{1}{l}{\cellcolor[rgb]{.988,.894,.839}0x000002FA0} & \multirow{2}{*}{\cellcolor[rgb]{1,1,1}\textbf{Unknown}} \\
     \multicolumn{3}{c}{\cellcolor[rgb]{ .851,  .882,  .949}\textbf{Bed\_To\_Toilet}} &  & \multicolumn{1}{l}{\cellcolor[rgb]{ .988,  .894,  .839}0000000000000000000000000000000001} & \multicolumn{1}{l}{\cellcolor[rgb]{ .988,  .894,  .839}0x000000001} & \\
\midrule   \multicolumn{3}{c}{} &     & \multicolumn{3}{c}{\cellcolor[rgb]{ .886,  .937,  .855}\textbf{Sleeping}} \\
\midrule   \rowcolor[rgb]{ .557,  .663,  .859} \multicolumn{3}{c}{\textbf{Sleeping}} & \cellcolor[rgb]{ 1,  1,  1} &\multicolumn{3}{l}{\cellcolor[rgb]{1,1,1}} \\
\midrule   \multicolumn{1}{l}{\cellcolor[rgb]{ 1,  .949,  .8} 0000000000000000000000000000000001} & \multicolumn{1}{l}{\cellcolor[rgb]{1,.949,.8} 0x000000001} & \multirow{3}{*}{\cellcolor[rgb]{1,1,1}\textbf{Unknown}} & \cellcolor[rgb]{ 1,  1,  1} & \multicolumn{3}{c}{\cellcolor[rgb]{ .663,  .816,  .557}\textbf{Bed\_To\_Toilet}} \\
    \multicolumn{1}{l}{\cellcolor[rgb]{1,.949,.8} 0000000000000000000010100100101101} & \multicolumn{1}{l}{\cellcolor[rgb]{ 1,.949,.8} 0x00000292D} &     & \cellcolor[rgb]{ 1,  1,  1} & \multicolumn{1}{l}{\cellcolor[rgb]{ .988,  .894,  .839}0000000000000000000000000000000001} & \multicolumn{1}{l}{\cellcolor[rgb]{.988,.894,.839}0x000000001} & \multirow{3}{*}{\cellcolor[rgb]{1,1,1}\textbf{Unknown}} \\
   \multicolumn{1}{l}{\cellcolor[rgb]{1,.949,.8} 0000000000000000000000000000000001} & \multicolumn{1}{l}{\cellcolor[rgb]{ 1,.949,.8} 0x000000001} &  \cellcolor[rgb]{1,1,1}    & \cellcolor[rgb]{ 1,  1,  1} & \multicolumn{1}{l}{\cellcolor[rgb]{ .988,  .894,  .839}0000000000000000000010111110100000} & \multicolumn{1}{l}{\cellcolor[rgb]{.988,.894,.839}0x000002FA0} &  \\
     \multicolumn{3}{c}{\cellcolor[rgb]{.851,.882,.949}\textbf{Bed\_To\_Toilet}} & \cellcolor[rgb]{1,1,1} & \multicolumn{1}{l}{\cellcolor[rgb]{.988,.894,.839}0000000000000000000010111110100000} & \multicolumn{1}{l}{\cellcolor[rgb]{ .988,  .894,  .839}0x000002FA0} &  \\
\midrule    \multicolumn{3}{c}{} &     & \multicolumn{3}{c}{\cellcolor[rgb]{.886,.937,.855}\textbf{Sleeping}} \\\midrule
    \rowcolor[rgb]{.557,.663,.859} \multicolumn{3}{c}{\textbf{Sleeping}} & \cellcolor[rgb]{1,1,1}&\multicolumn{3}{l}{\cellcolor[rgb]{1,1,1}} \\\midrule
    \multicolumn{1}{l}{\cellcolor[rgb]{ 1,  .949,  .8} 0000000000000000000000000000000001} & \multicolumn{1}{l}{\cellcolor[rgb]{1,.949,.8}0x000000001} & \multirow{6}{*}{\cellcolor[rgb]{1,1,1}\textbf{Unknown}} & \cellcolor[rgb]{1,1,1} & \multicolumn{3}{c}{\cellcolor[rgb]{ .663,  .816,  .557}\textbf{Bed\_To\_Toilet}} \\
    \multicolumn{1}{l}{\cellcolor[rgb]{ 1,  .949,  .8}0000000000000000000000000000000001} & \multicolumn{1}{l}{\cellcolor[rgb]{1,.949,.8}0x000000001} &     & \cellcolor[rgb]{1,1,1} & \multicolumn{1}{l}{\cellcolor[rgb]{ .988,  .894,  .839}0000000000000000000011010001011001} & \multicolumn{1}{l}{\cellcolor[rgb]{ .988,  .894,  .839}0x000003459} & \multirow{2}{*}{\cellcolor[rgb]{ 1,  1,  1}\textbf{Unknown}} \\
    \multicolumn{1}{l}{\cellcolor[rgb]{ 1,  .949,  .8}0000000000000000000000000000000001} & \multicolumn{1}{l}{\cellcolor[rgb]{ 1,  .949,  .8}0x000000001} &     & \cellcolor[rgb]{ 1,  1,  1} & \multicolumn{1}{l}{\cellcolor[rgb]{ .988,  .894,  .839}0000000000000000000011010001011001} & \multicolumn{1}{l}{\cellcolor[rgb]{ .988,  .894,  .839}0x000003459} &  \\
     \multicolumn{1}{l}{\cellcolor[rgb]{ 1,  .949,  .8}0000000000000000000010100100101101} & \multicolumn{1}{l}{\cellcolor[rgb]{ 1,  .949,  .8}0x00000292D} &     & \cellcolor[rgb]{ 1,  1,  1} & \multicolumn{3}{c}{\cellcolor[rgb]{ .886,  .937,  .855}\textbf{Sleeping}} \\
    \multicolumn{1}{l}{\cellcolor[rgb]{ 1,  .949,  .8}0000000000000000000000000000000001} & \multicolumn{1}{l}{\cellcolor[rgb]{ 1,  .949,  .8}0x000000001} &     & \cellcolor[rgb]{ 1,  1,  1} &\multicolumn{3}{l}{\cellcolor[rgb]{1,1,1}} \\
   \multicolumn{1}{l}{\cellcolor[rgb]{ 1,  .949,  .8}0000000000000000000010111110100000} & \multicolumn{1}{l}{\cellcolor[rgb]{ 1,  .949,  .8}0x000002FA0} &     & \cellcolor[rgb]{ 1,  1,  1} & \multicolumn{3}{c}{\cellcolor[rgb]{ .663,  .816,  .557}\textbf{Bed\_To\_Toilet}} \\
    \rowcolor[rgb]{ .851,  .882,  .949} \multicolumn{3}{c}{\textbf{Bed\_To\_Toilet}} & \cellcolor[rgb]{ 1,  1,  1} & \multicolumn{1}{l}{\cellcolor[rgb]{ .988,  .894,  .839}0000000000000000000011010001011001} & \multicolumn{1}{l}{\cellcolor[rgb]{ .988,  .894,  .839}0x000003459} & \multirow{2}{*}{\cellcolor[rgb]{ 1,  1,  1}\textbf{Unknown}} \\
     \multicolumn{3}{c}{} &     & \multicolumn{1}{l}{\cellcolor[rgb]{ .988,  .894,  .839}0000000000000000000000000000000001} & \multicolumn{1}{l}{\cellcolor[rgb]{ .988,  .894,  .839}0x000000001} &  \\\midrule
     \rowcolor[rgb]{ .557,  .663,  .859} \multicolumn{3}{c}{\textbf{Sleeping}} & \cellcolor[rgb]{ 1,  1,  1} & \multicolumn{3}{c}{\cellcolor[rgb]{ .886,  .937,  .855}\textbf{Sleeping}} \\\midrule
     \multicolumn{1}{l}{\cellcolor[rgb]{ 1,  .949,  .8}0000000000000000000010100100101101} & \multicolumn{1}{l}{\cellcolor[rgb]{ 1,  .949,  .8}0x00000292D} & \multirow{6}{*}{\cellcolor[rgb]{ 1,  1,  1}\textbf{Unknown}}& &\multicolumn{3}{l}{\cellcolor[rgb]{1,1,1}} \\
    \multicolumn{1}{l}{\cellcolor[rgb]{ 1,  .949,  .8}0000000000000000000010100100101101} & \multicolumn{1}{l}{\cellcolor[rgb]{ 1,  .949,  .8}0x00000292D} &     & \cellcolor[rgb]{ 1,  1,  1} & \multicolumn{3}{c}{\cellcolor[rgb]{ .663,  .816,  .557}\textbf{Bed\_To\_Toilet}} \\
    \multicolumn{1}{l}{\cellcolor[rgb]{ 1,  .949,  .8}0000000000000000000000000000000001} & \multicolumn{1}{l}{\cellcolor[rgb]{ 1,  .949,  .8}0x000000001} &     & \cellcolor[rgb]{ 1,  1,  1} & \multicolumn{1}{l}{\cellcolor[rgb]{ .988,  .894,  .839}0000000000000000000000000000000001} & \multicolumn{1}{l}{\cellcolor[rgb]{ .988,  .894,  .839}0x000000001} & \multirow{4}{*}{\cellcolor[rgb]{ 1,  1,  1}\textbf{Unknown}} \\
    \multicolumn{1}{l}{\cellcolor[rgb]{ 1,  .949,  .8}0000000000000000000000000000000001} & \multicolumn{1}{l}{\cellcolor[rgb]{ 1,  .949,  .8}0x000000001} &     & \cellcolor[rgb]{ 1,  1,  1} & \multicolumn{1}{l}{\cellcolor[rgb]{ .988,  .894,  .839}0000000000000000000000000000000001} & \multicolumn{1}{l}{\cellcolor[rgb]{ .988,  .894,  .839}0x000000001}  &\\
    \multicolumn{1}{l}{\cellcolor[rgb]{ 1,  .949,  .8}0000000000000000000000000000000001} & \multicolumn{1}{l}{\cellcolor[rgb]{ 1,  .949,  .8}0x000000001} &     & \cellcolor[rgb]{ 1,  1,  1} & \multicolumn{1}{l}{\cellcolor[rgb]{ .988,  .894,  .839}0000000000000000000011010001011001} & \multicolumn{1}{l}{\cellcolor[rgb]{ .988,  .894,  .839}0x000003459} &   \\
    \multicolumn{1}{l}{\cellcolor[rgb]{ 1,  .949,  .8}0000000000000000000000000000000001} & \multicolumn{1}{l}{\cellcolor[rgb]{ 1,  .949,  .8}0x000000001} &     & \cellcolor[rgb]{ 1,  1,  1} & \multicolumn{1}{l}{\cellcolor[rgb]{ .988,  .894,  .839}0000000000000000000000000000000001} & \multicolumn{1}{l}{\cellcolor[rgb]{ .988,  .894,  .839}0x000000001} & \cellcolor[rgb]{1,1,1}  \\
\midrule    \rowcolor[rgb]{ .851,  .882,  .949} \multicolumn{3}{c}{\textbf{Bed\_To\_Toilet}} & \cellcolor[rgb]{ 1,  1,  1} & \multicolumn{3}{c}{\cellcolor[rgb]{ .886,  .937,  .855}\textbf{Sleeping}} \\
\bottomrule    \end{tabular}}%
  \label{tab:addlabel}%
\end{table} 

\textls[-20]{Inspired by this argument, we treat the gaps in this paradigm as distinguishing signatures of the encapsulating activities. This kind of gap handling may help identification of the encapsulating activities, since successful identification of such gaps guides identification of the encapsulating activities.}

Theoretically, the number of interactivity labels equals square of the number of activities. However, practically, the matrix of the interactivity labels is sparse. This sparsity nature is inherited from the zero-probability links of the HMM state diagram. This means that the added states to the state diagram will not necessarily equal $N^2$.

\begin{figure}
  \centering
 \includegraphics[width=.85\textwidth]{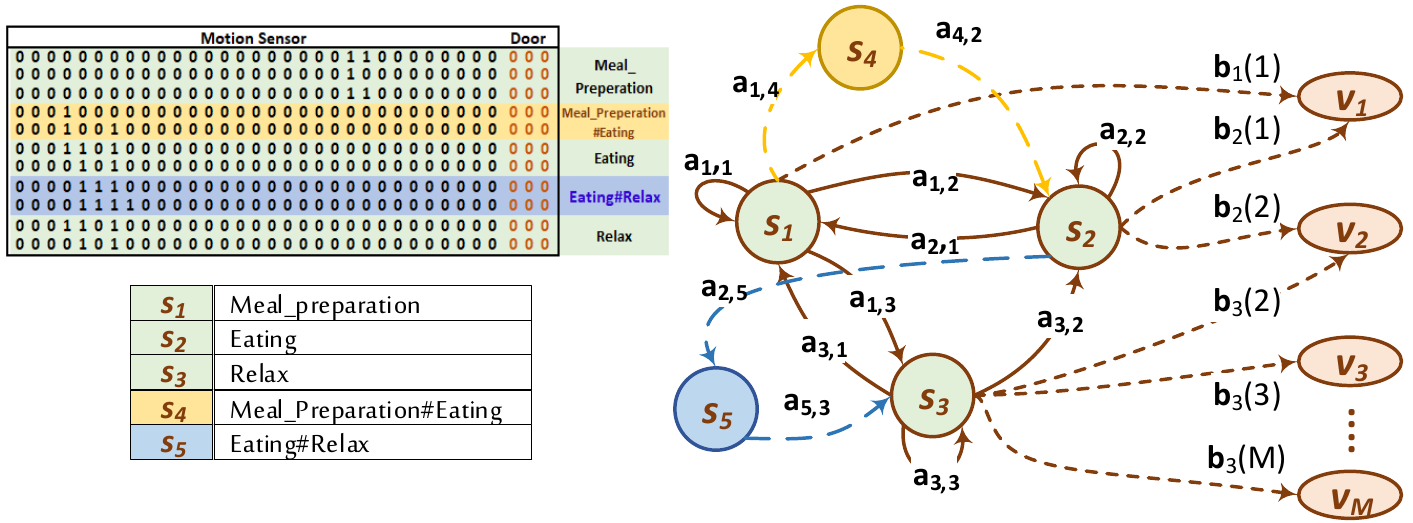}
\caption{Paradigm~\#3: A new state is created for each annotation gap and stamped by its ordered encapsulating activities in order.}
\label{fig:All4}
\end{figure}

This procedure leads to reformulation of the HMM model parameters by increasing the cardinality if the state space, as follows:
\begin{equation}
    |S_{p3}| = N+N^*;\;N^*\leq N^2
\end{equation}

\noindent where $S_{p3}$ is the state space of paradigm~\#3. Therefore, the state space cardinality of this paradigm is increased from $N$ in paradigm~\#1 and from $N+1$ in paradigm~\#2. All consequent transition and emission models are re-estimated accordingly. Figure~\ref{fig:All4} shows a graphical illustration of the operation of this paradigm.

% ...

%



\section{Experimental Setup}
\label{sec:setup}
The main objective of this work is to find the best way to handle annotation gaps in the raw input human activity data. We set up our evaluation experimentation to highlight the effect of each of the proposed three paradigms on different activity recognition metrics. In Section~\ref{sec:dataset}, the used dataset for evaluation are described. In Section~\ref{sec:eval_metrics}, we illustrate the procedures, which we followed in the evaluation process and the used metrics.

\subsection{Dataset}
\label{sec:dataset}

\begin{figure}
  \centering
    \includegraphics[width=0.8\textwidth]{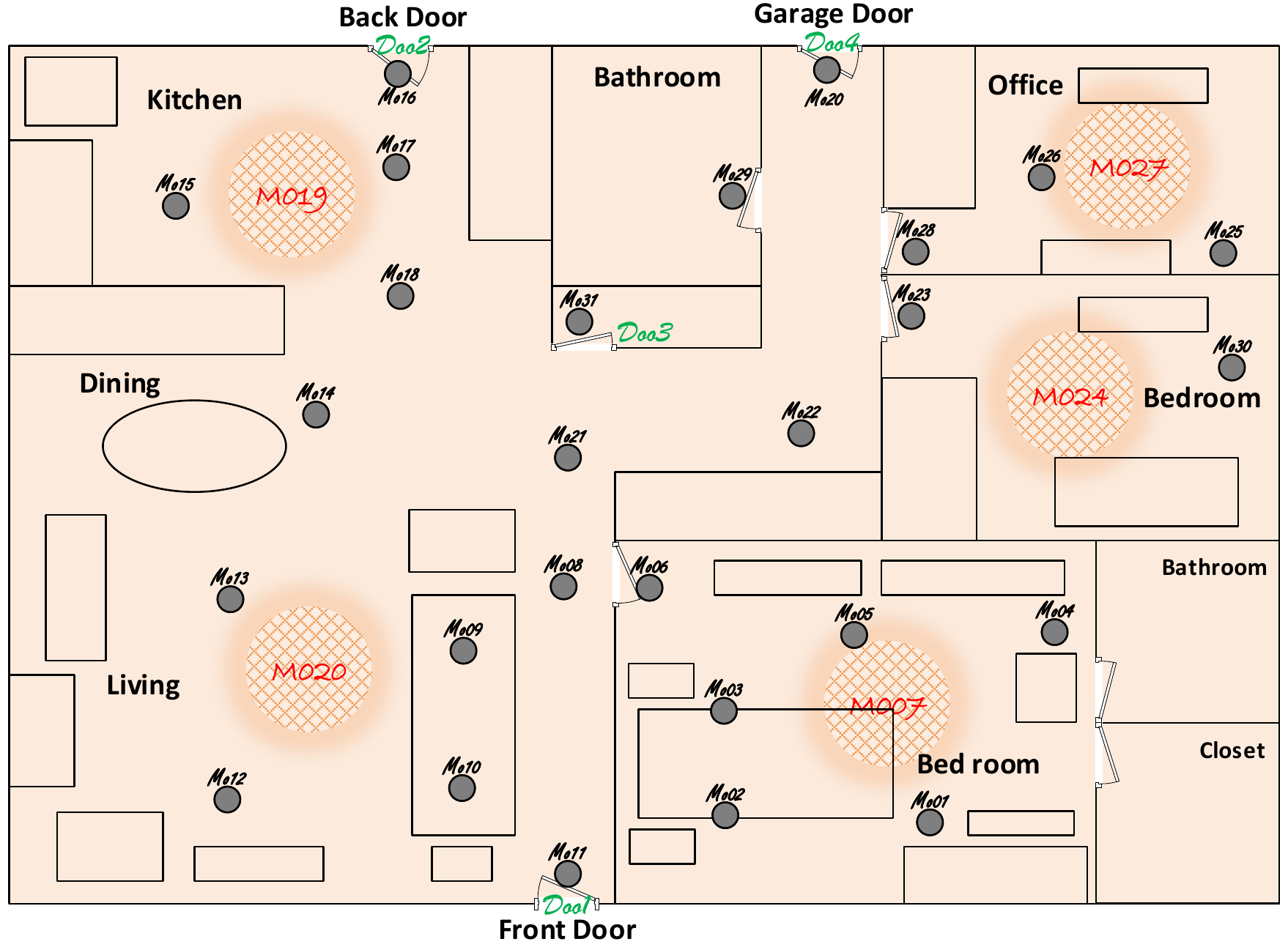}
\caption{Sensor placement inside the experimental environment~\cite{Aruba}.}
\label{fig:aruba_setup}
\end{figure}

\textls[-15]{The proposed method is applied on a dataset collected from the WSU-CASAS smart home project~\cite{Aruba}. The set consists of more than 2.5 M sensor readings for a single female household for the whole day (24 h) over more than nine months. Thirty-four sensors are distributed all over the house. There are $31$ binary motion sensors plus $3$ door-closure binary sensors. The placement of these sensors inside the house is shown in Figure~\ref{fig:aruba_setup}. In addition to sensor readings and timestamps, manual annotation is carried out to the data to label current household's activities by its ground truth labels. There are nine ground truth labels representing nine activities. For evaluation, we use the annotation gaps that already exist in the dataset. We do not create our own annotation gaps. Therefore, the annotation gaps used for evaluation is completely unbiased and reflect typical realistic annotation behavior.}

\textls[-20]{The readings of the sensors are represented by a $34$ bits binary code $V^i$, where $i$ is the observation instance. Some examples explaining the structure of these codes are shown in Table~\ref{tab:sensor_code}. Bits from $v^i_0$ to $v^i_{30}$ represent the $31$ motion sensors while the bits in the range $v^i_{31}-v^i_{33}$ denote the $3$ door-closure sensors.}

\begin{table}
  \centering
  \caption{Exemplar  codes  of  sensor  readings.   Thirty-one  bits  are  assigned by motion sensors, and three bits by door-closure sensors.}
    \scalebox{.7}[.7]{\begin{tabular}{crrrrrrrrrrrrrrrrrrrrrrrrrrrrrrrrrrrrc}

   \toprule
    \multicolumn{31}{c}{\footnotesize{\textbf{Motion Sensor}}}&      \multicolumn{4}{c}{\footnotesize{\textbf{Door}}}  \\
   \toprule
     \footnotesize{0} & \footnotesize{0} & \footnotesize{0} & \footnotesize{0} & \footnotesize{0} & \footnotesize{0} & \footnotesize{0} & \footnotesize{0} & \footnotesize{0} & \footnotesize{0} & \footnotesize{0} & \footnotesize{0} & \footnotesize{0} & \footnotesize{0} & \footnotesize{0} & \footnotesize{0} & \footnotesize{0} & \footnotesize{0} & \footnotesize{0} & \footnotesize{0} & \footnotesize{0} & \footnotesize{1} & \footnotesize{1} & \footnotesize{0} & \footnotesize{0} & \footnotesize{0} & \footnotesize{0} & \footnotesize{0} & \footnotesize{0} & \footnotesize{0} & \footnotesize{0} &  \textcolor[rgb]{ .776,  .349,  .067}{~}  & \cellcolor[rgb]{ .859,  .859,  .859}\textcolor[rgb]{ .776,  .349,  .067}{\footnotesize{0}} & \cellcolor[rgb]{ .859,  .859,  .859}\textcolor[rgb]{ .776,  .349,  .067}{\footnotesize{1}} & \cellcolor[rgb]{ .859,  .859,  .859}\textcolor[rgb]{ .776,  .349,  .067}{\footnotesize{0}}    \\
    
         \footnotesize{0} & \footnotesize{0} & \footnotesize{0} & \footnotesize{0} & \footnotesize{0} & \footnotesize{0} & \footnotesize{0} & \footnotesize{0} & \footnotesize{0} & \footnotesize{0} & \footnotesize{0} & \footnotesize{0} & \footnotesize{0} & \footnotesize{0} & \footnotesize{0} & \footnotesize{0} & \footnotesize{0} & \footnotesize{0} & \footnotesize{0} & \footnotesize{0} & \footnotesize{0} & \footnotesize{1} & \footnotesize{0} & \footnotesize{0} & \footnotesize{0} & \footnotesize{0} & \footnotesize{0} & \footnotesize{0} & \footnotesize{0} & \footnotesize{0} & \footnotesize{0} &     & \cellcolor[rgb]{ .859,  .859,  .859}\textcolor[rgb]{ .776,  .349,  .067}{\footnotesize{1}} & \cellcolor[rgb]{ .859,  .859,  .859}\textcolor[rgb]{ .776,  .349,  .067}{\footnotesize{1}} & \cellcolor[rgb]{ .859,  .859,  .859}\textcolor[rgb]{ .776,  .349,  .067}{\footnotesize{0}}\\
         \footnotesize{0} & \footnotesize{0} & \footnotesize{0} & \footnotesize{1} & \footnotesize{0} & \footnotesize{0} & \footnotesize{1} & \footnotesize{0} & \footnotesize{0} & \footnotesize{0} & \footnotesize{0} & \footnotesize{0} & \footnotesize{0} & \footnotesize{0} & \footnotesize{0} & \footnotesize{0} & \footnotesize{0} & \footnotesize{0} & \footnotesize{0} & \footnotesize{0} & \footnotesize{0} & \footnotesize{0} & \footnotesize{0} & \footnotesize{0} & \footnotesize{0} & \footnotesize{0} & \footnotesize{0} & \footnotesize{0} & \footnotesize{0} & \footnotesize{0} & \footnotesize{0} &     & \cellcolor[rgb]{ .859,  .859,  .859}\textcolor[rgb]{ .776,  .349,  .067}{\footnotesize{0}} & \cellcolor[rgb]{ .859,  .859,  .859}\textcolor[rgb]{ .776,  .349,  .067}{\footnotesize{0}} & \cellcolor[rgb]{ .859,  .859,  .859}\textcolor[rgb]{ .776,  .349,  .067}{\footnotesize{1}} \\
         \footnotesize{0} & \footnotesize{0} & \footnotesize{0} & \footnotesize{1} & \footnotesize{0} & \footnotesize{0} & \footnotesize{0} & \footnotesize{0} & \footnotesize{0} & \footnotesize{0} & \footnotesize{0} & \footnotesize{0} & \footnotesize{0} & \footnotesize{0} & \footnotesize{0} & \footnotesize{0} & \footnotesize{0} & \footnotesize{0} & \footnotesize{0} & \footnotesize{0} & \footnotesize{0} & \footnotesize{0} & \footnotesize{0} & \footnotesize{0} & \footnotesize{0} & \footnotesize{0} & \footnotesize{0} & \footnotesize{0} & \footnotesize{0} & \footnotesize{0} & \footnotesize{0} &     & \cellcolor[rgb]{ .859,  .859,  .859}\textcolor[rgb]{ .776,  .349,  .067}{\footnotesize{1}} & \cellcolor[rgb]{ .859,  .859,  .859}\textcolor[rgb]{ .776,  .349,  .067}{\footnotesize{1}} & \cellcolor[rgb]{ .859,  .859,  .859}\textcolor[rgb]{ .776,  .349,  .067}{\footnotesize{1}} \\
         \footnotesize{0} & \footnotesize{0} & \footnotesize{0} & \footnotesize{1} & \footnotesize{0} & \footnotesize{0} & \footnotesize{1} & \footnotesize{0} & \footnotesize{0} & \footnotesize{0} & \footnotesize{0} & \footnotesize{0} & \footnotesize{0} & \footnotesize{0} & \footnotesize{0} & \footnotesize{0} & \footnotesize{0} & \footnotesize{0} & \footnotesize{0} & \footnotesize{0} & \footnotesize{0} & \footnotesize{0} & \footnotesize{0} & \footnotesize{0} & \footnotesize{0} & \footnotesize{0} & \footnotesize{0} & \footnotesize{0} & \footnotesize{0} & \footnotesize{0} & \footnotesize{0} &     & \cellcolor[rgb]{ .859,  .859,  .859}\textcolor[rgb]{ .776,  .349,  .067}{\footnotesize{0}} & \cellcolor[rgb]{ .859,  .859,  .859}\textcolor[rgb]{ .776,  .349,  .067}{\footnotesize{0}} & \cellcolor[rgb]{ .859,  .859,  .859}\textcolor[rgb]{ .776,  .349,  .067}{\footnotesize{0}} \\
         \footnotesize{0} & \footnotesize{0} & \footnotesize{0} & \footnotesize{1} & \footnotesize{1} & \footnotesize{0} & \footnotesize{1} & \footnotesize{0} & \footnotesize{0} & \footnotesize{0} & \footnotesize{0} & \footnotesize{0} & \footnotesize{0} & \footnotesize{0} & \footnotesize{0} & \footnotesize{0} & \footnotesize{0} & \footnotesize{0} & \footnotesize{0} & \footnotesize{0} & \footnotesize{0} & \footnotesize{0} & \footnotesize{0} & \footnotesize{0} & \footnotesize{0} & \footnotesize{0} & \footnotesize{0} & \footnotesize{0} & \footnotesize{0} & \footnotesize{0} & \footnotesize{0} &     & \cellcolor[rgb]{ .859,  .859,  .859}\textcolor[rgb]{ .776,  .349,  .067}{\footnotesize{0}} & \cellcolor[rgb]{ .859,  .859,  .859}\textcolor[rgb]{ .776,  .349,  .067}{\footnotesize{0}} & \cellcolor[rgb]{ .859,  .859,  .859}\textcolor[rgb]{ .776,  .349,  .067}{\footnotesize{0}} \\
         \footnotesize{0} & \footnotesize{0} & \footnotesize{0} & \footnotesize{0} & \footnotesize{1} & \footnotesize{0} & \footnotesize{1} & \footnotesize{0} & \footnotesize{0} & \footnotesize{0} & \footnotesize{0} & \footnotesize{0} & \footnotesize{0} & \footnotesize{0} & \footnotesize{0} & \footnotesize{0} & \footnotesize{0} & \footnotesize{0} & \footnotesize{0} & \footnotesize{0} & \footnotesize{0} & \footnotesize{0} & \footnotesize{0} & \footnotesize{0} & \footnotesize{0} & \footnotesize{0} & \footnotesize{0} & \footnotesize{0} & \footnotesize{0} & \footnotesize{0} & \footnotesize{0} &     & \cellcolor[rgb]{ .859,  .859,  .859}\textcolor[rgb]{ .776,  .349,  .067}{\footnotesize{0}} & \cellcolor[rgb]{ .859,  .859,  .859}\textcolor[rgb]{ .776,  .349,  .067}{\footnotesize{1}} & \cellcolor[rgb]{ .859,  .859,  .859}\textcolor[rgb]{ .776,  .349,  .067}{\footnotesize{0}}  \\
         \footnotesize{0} & \footnotesize{0} & \footnotesize{0} & \footnotesize{0} & \footnotesize{1} & \footnotesize{1} & \footnotesize{1} & \footnotesize{0} & \footnotesize{0} & \footnotesize{0} & \footnotesize{0} & \footnotesize{0} & \footnotesize{0} & \footnotesize{0} & \footnotesize{0} & \footnotesize{0} & \footnotesize{0} & \footnotesize{0} & \footnotesize{0} & \footnotesize{0} & \footnotesize{0} & \footnotesize{0} & \footnotesize{0} & \footnotesize{0} & \footnotesize{0} & \footnotesize{0} & \footnotesize{0} & \footnotesize{0} & \footnotesize{0} & \footnotesize{0} & \footnotesize{0} &     & \cellcolor[rgb]{ .859,  .859,  .859}\textcolor[rgb]{ .776,  .349,  .067}{\footnotesize{1}} & \cellcolor[rgb]{ .859,  .859,  .859}\textcolor[rgb]{ .776,  .349,  .067}{\footnotesize{1}} & \cellcolor[rgb]{ .859,  .859,  .859}\textcolor[rgb]{ .776,  .349,  .067}{\footnotesize{0}}    \\
         \footnotesize{0} & \footnotesize{0} & \footnotesize{0} & \footnotesize{0} & \footnotesize{1} & \footnotesize{1} & \footnotesize{1} & \footnotesize{1} & \footnotesize{0} & \footnotesize{0} & \footnotesize{0} & \footnotesize{0} & \footnotesize{0} & \footnotesize{0} & \footnotesize{0} & \footnotesize{0} & \footnotesize{0} & \footnotesize{0} & \footnotesize{0} & \footnotesize{0} & \footnotesize{0} & \footnotesize{0} & \footnotesize{0} & \footnotesize{0} & \footnotesize{0} & \footnotesize{0} & \footnotesize{0} & \footnotesize{0} & \footnotesize{0} & \footnotesize{0} & \footnotesize{0} &     & \cellcolor[rgb]{ .859,  .859,  .859}\textcolor[rgb]{ .776,  .349,  .067}{\footnotesize{0}} & \cellcolor[rgb]{ .859,  .859,  .859}\textcolor[rgb]{ .776,  .349,  .067}{\footnotesize{0}} & \cellcolor[rgb]{ .859,  .859,  .859}\textcolor[rgb]{ .776,  .349,  .067}{\footnotesize{0}}    \\
   \hline
    \end{tabular}}%
  \label{tab:sensor_code}%
\end{table}




\subsection{Evaluation Procedures and Metrics}
\label{sec:eval_metrics}

\textls[-15]{As illustrated in Section~\ref{sec:dataset}, the input sensor dataset is semi-annotated manually. The main objective of an activity recognition system is to achieve high recognition rates. The recognition performance is affected by different factors. In HMM-based models, the succession of states and the length of each state directly influence both the transition and emission models. This gives special importance to the length of the sampling time interval, $\Delta t$. In particular, if the sampling time interval is much longer than the activity duration, some activities may be lost and not reflected to the transition model. On the other side, if it is too small, repetition overhead will be increased during the HMM model estimation affecting the state transition distribution. Therefore, a tradeoff is needed. However, the recall-precision pairs are highly dependent on both $\Delta t$ and the activities' duration. This means that changing $\Delta t$ to obtain a generalized stable recall-precision analysis such as ROC that is globally applicable to any activity pattern is questionable. Therefore, reaching a general recommendation of the best sampling interval in terms of the intervals of the sensor readings is the best investigation outcome rather than building a conclusion on an activity pattern-dependent outcome. Therefore, we opted to investigate the most proper value of $\Delta t$ independently of the underlying activity duration. For the used dataset, the average activity interval between sensor readings is about 11 seconds. We tested different values for $\Delta t$ and found that an interval of 7 seconds gives a good basis for all paradigms. Therefore, we recommend using  $\Delta t $ values between $50\%$ and $65\%$ of the average sensor inter-reading intervals. This sampling rate guarantees a relatively ``succinct'' coverage of almost all the recorded readings.}

We resampled the input sensor readings using $\Delta t = 7$ s. Then, we applied the Viterbi algorithm to estimate the emission and transition models. The size of the generated samples exceeded 2 M samples. 60\% of these samples were used as the training dataset for model estimation. The remaining 40\% were used to evaluate the performance.

Since the main investigation target of the proposed paradigms is the activity recognition performance, we considered four major recognition evaluation metrics. Specifically, we used \textit{recall, accuracy, precision,} and \textit{specificity} metrics for this purpose.

\textit{Recall} measures the fraction of the correctly recognized instances of an activity to the total number of the retrieved instances of this specific activity, as shown in the following equation:
\begin{equation}
    Recall = \frac{TP}{TP+FN}
\end{equation}

\noindent where $TP$ equals the number of true positives and $FN$ equals the number of false negatives.

The \textit{Precision} metric is used to quantify the positive predictive values of recognizing an activity.
\begin{equation}
    Precision = \frac{TP}{TP+FP}
\end{equation}

\textls[-20]{The closeness of the total retrieved activities to the true recognition values is measured by the \textit{Accuracy}.}
\begin{equation}
    Accuracy = \frac{TP+TN}{TP+TN+FP+FN}
\end{equation}

\noindent where $TN$ and $FP$ are the number of true negatives and false positives, respectively.

Finally, the True Negative Rate (TNR), or \textit{Specificity}, quantifies how specific the classifier is, with~respect to the other activities. In other words, how much the activity does not influence other activities when they are classified.
\begin{equation}
    Specificity = \frac{TN}{TN+FP}
\end{equation}


\section{Results and Discussion}\label{sec:discussion}
Handling the annotation gaps (unlabeled data) in the raw input human activity data is a challenging task. We proposed three paradigms from different perspectives to investigate the best way to handle this unlabeled data. To evaluate and quantify the performance of the proposed paradigms, we used the aforementioned four metrics with the evaluation dataset for every single activity.

\begin{figure}
  \centering
 \includegraphics[width=0.8\textwidth]{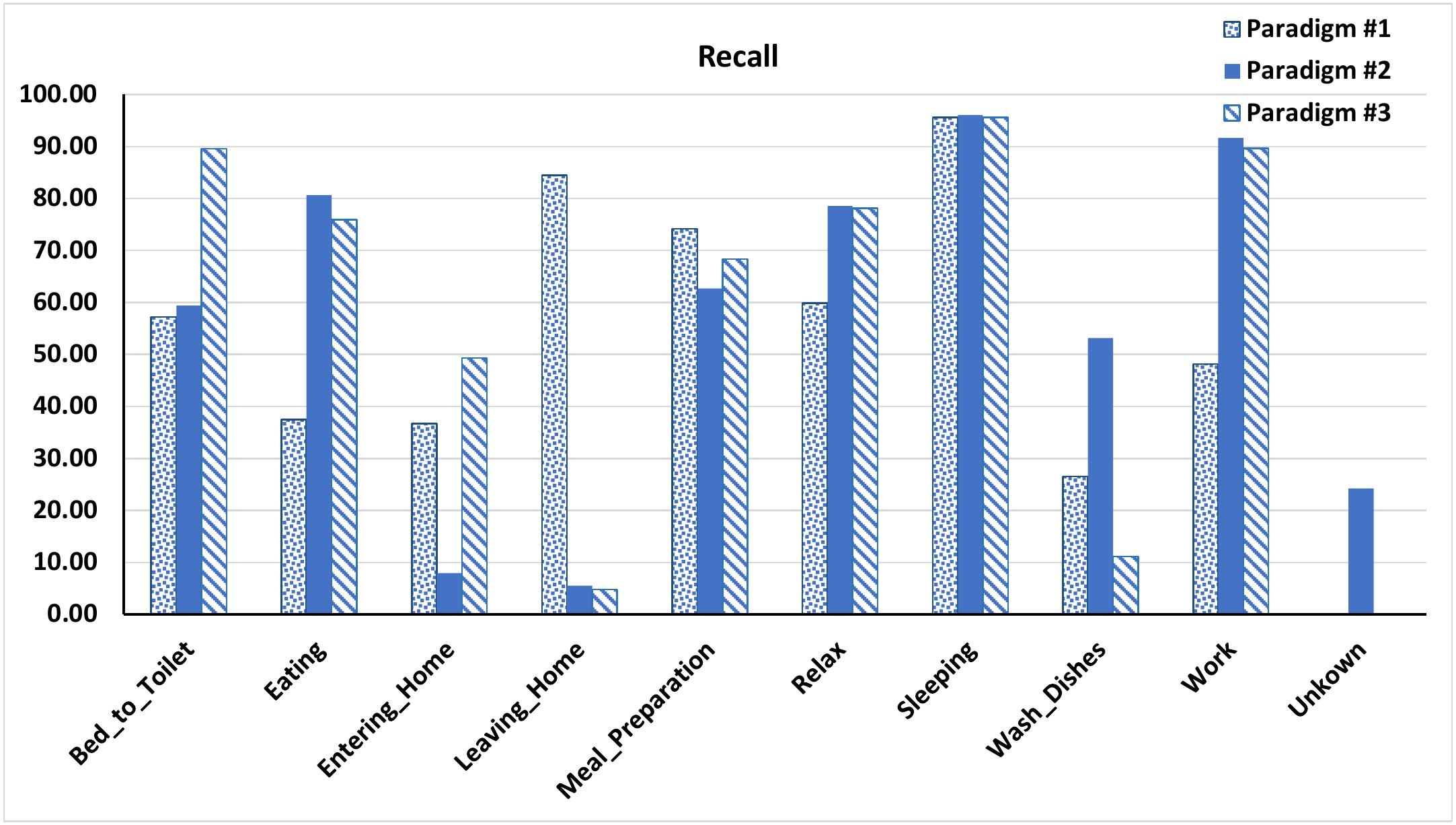}
\caption{Recall values of the proposed three paradigms.}
\label{fig:Recall}
\end{figure}

Figure~\ref{fig:Recall} shows the recall values. As for paradigm~\#1, removing all annotation gaps, it shows high recall values in three activities: \textit{Leaving-Home}, \textit{Sleeping}, and \textit{Meal-Preparation}. For the \textit{Meal-preparation} and \textit{Sleeping} activities, the differences between the recall values of the three paradigms are not so big. However, for the \textit{Leaving-Home} activity, the superiority of paradigm~\#1 is obvious. In fact, this~is a logical behavior. In particular, if we think about the core operation of this paradigm, we see that the interactivity annotation after the \textit{Leaving-Home} activity is usually removed. This is because the human annotator does not have an `out-of-home'-like label. Therefore, he/she usually leaves all input readings between the \textit{Leaving-Home} and the consequent activity, which is \textit{Entering-Home} blank. Paradigm~\#1 by nature omits these gaps resulting in a `certain', yet logical, sorted order correlation between the \textit{Leaving-Home} and the \textit{Entering-Home} activities. In contrast to the other two paradigms, they give some label to the in-between  gaps separating these two activities. Since the used dataset is for a single household, ideally there should be no readings between these two activities except sensors noise. The given label for this gap, either it is an \textit{Unknown} label as in paradigm~\#2 or \textit{Leaving-Entering} as in paradigm~\#3, easily gets confused with other `idle-similar' activities. This argument is supported by investigating the top matches to the \textit{Leaving-Home} activity under the operation of the two paradigms. The top classification of the \textit{Leaning-Home} activity under paradigms~\#2 and~\#3 is \textit{Sleeping}. The \textit{Sleeping} activity relatively conforms with the gap after \textit{Leaving-Home} property of being idle till the beginning of the following activity.
To overcome this drawback, we suggest to extend the annotation of such kinds of activities to the successive one; or to create a new dedicated label for these specific gaps, e.g., \textit{Out-of-Home}, as shown later in this section.

Paradigm~\#2 does not have a bold improvement except for activities which have low repeatability of their successor activities, e.g., \textit{Wash-Dishes}.

\begin{figure}
  \centering
 \includegraphics[width=0.8\textwidth]{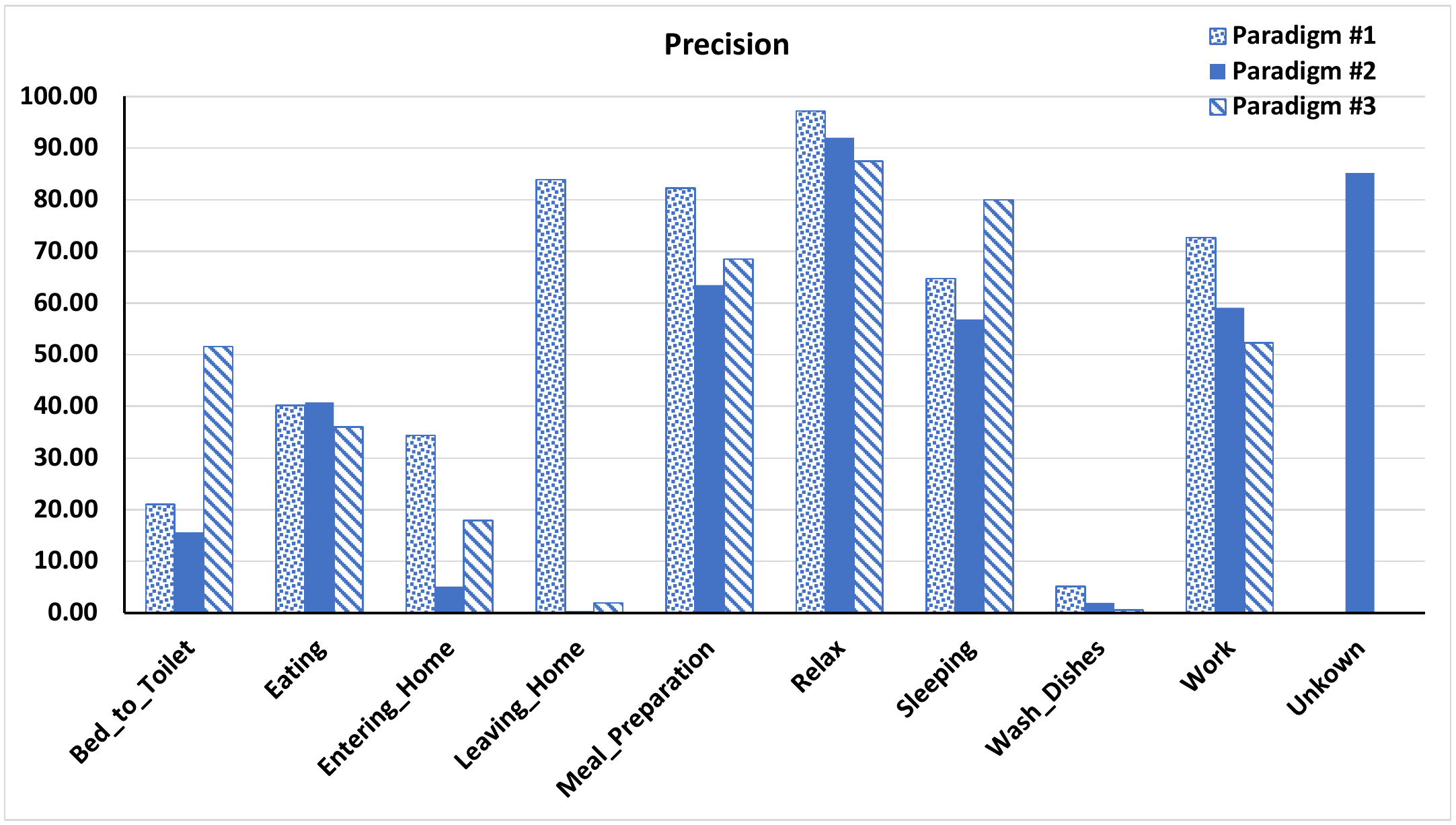}
\caption{Precision evaluation of the proposed paradigms.}
\label{fig:Precision}
\end{figure}

The precision values show similar behavior to recall's with slight differences, as shown in Figure~\ref{fig:Precision}. Therefore, the same preceding discussion applies to this measure.

\begin{figure}
  \centering
 \includegraphics[width=0.8\textwidth]{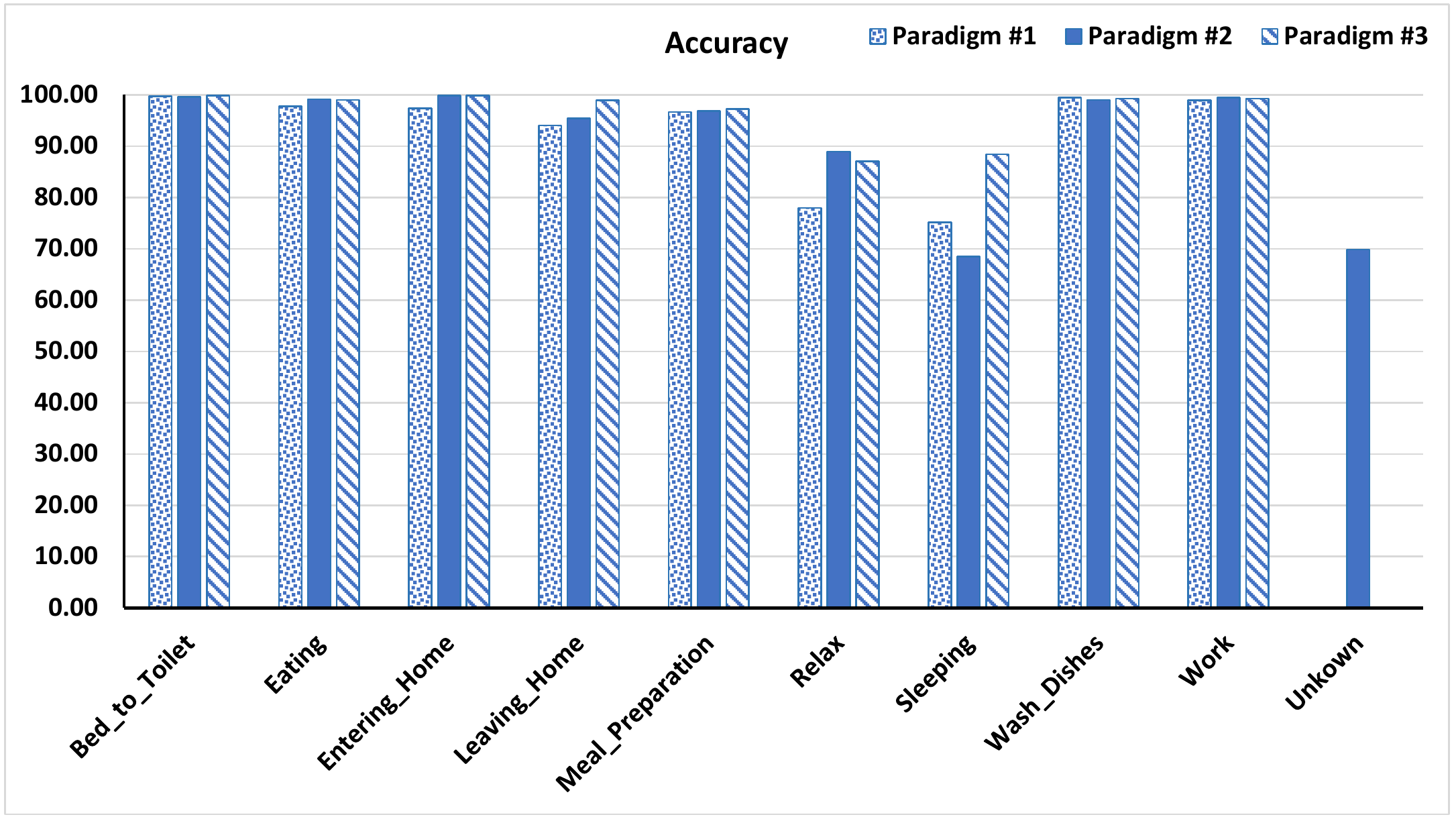}
\caption{Accuracy values of the proposed paradigms. Note the improvement of \textit{Sleeping} recognition with paradigm~\#3, which characterizes the encapsulating activities. This achieves some degree of distinction from other idle-tended activities such as \textit{Relax}.}
\label{fig:Accuracy}
\end{figure}

\begin{figure}
  \centering
 \includegraphics[width=0.8\textwidth]{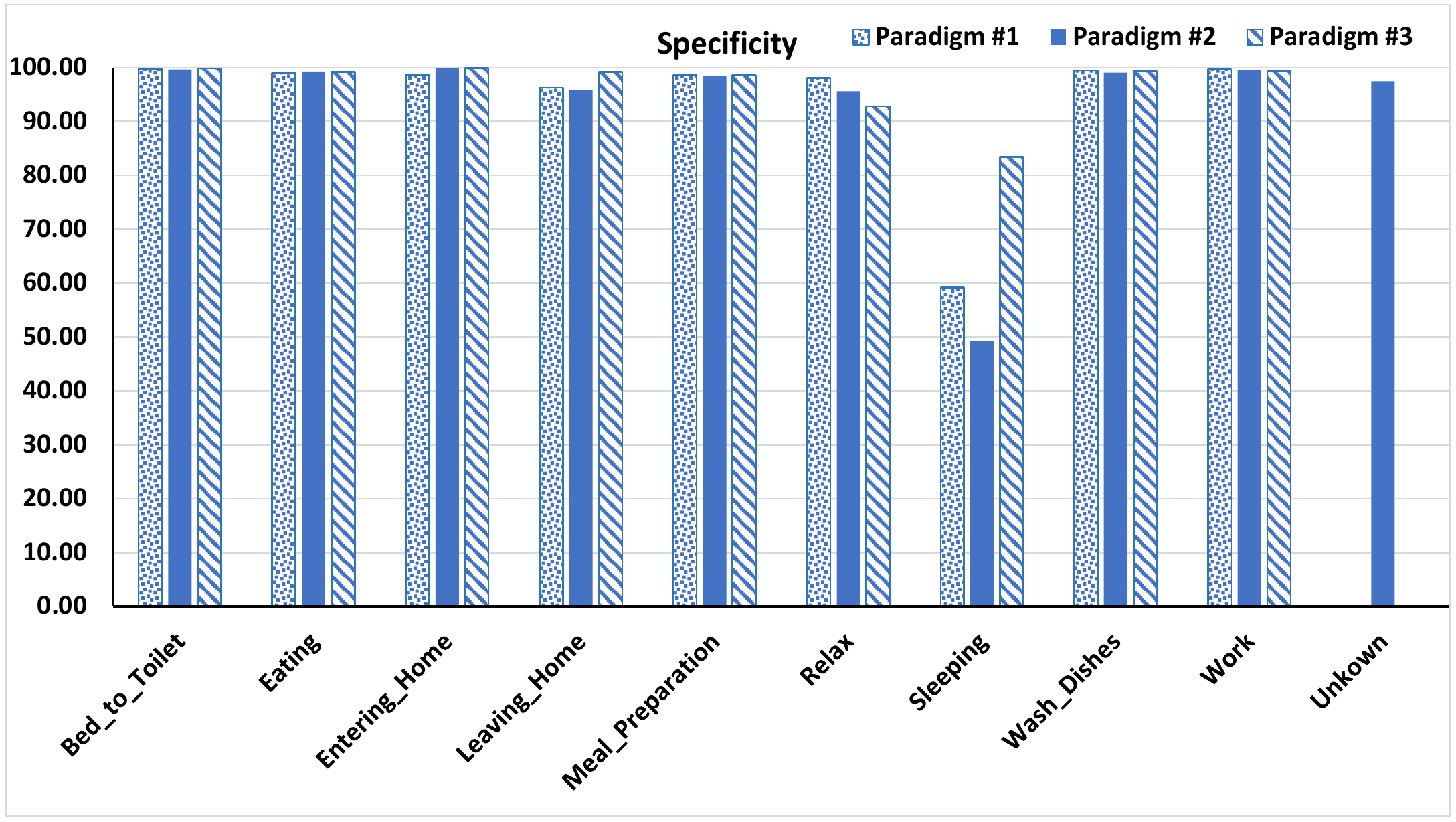}
\caption{Specificity values of the proposed three paradigms.}
\label{fig:Specificity}
\end{figure}

For the accuracy and specificity metrics shown in Figures~\ref{fig:Accuracy} and~\ref{fig:Specificity} respectively, all paradigms show high performance in slight superiority of paradigm~\#3. The only case in which paradigm~\#3 achieves noticeable improvement is with the \textit{Sleeping} activity. This is because of the repeatability nature of the successor activity, which is the \textit{Bed-to-Toilette} activity. Giving a unique label to the intermediate input between these two activities supports correct recognition decision.

\unskip

\textls[-15]{We developed an auxiliary preprocessing procedure to aid the model estimation and classification of HMM. Some erroneously missing annotations can be obviously restored. For example, as mentioned earlier, the annotation gap between activities such as \textit{Leaving\_Home} and \textit{Entering\_Home} is not logic, since it is obviously known that any recorded sensor readings between these two activities is erroneous. Therefore, the two activities should be adjacently connected. It is expected that eliminating this gap by extending the preceding \textit{Leaving\_Home} activity is hypothetically better for recognition performance. Activity extension in this specific case may be better than complete gap removal, as extension will reflect practical activity duration. Paradigm~\#3 blindly handles all such cases by giving these kinds of gaps unique labels. Nonetheless even if this gap is assigned a unique label, it will be very similar to idle-tended activities, e.g., \textit{Sleeping} and \textit{Relax}. When this semantic preprocessing step is performed, a~hybrid combination of paradigm~\#3 and paradigm~\#1, with slight modification by extension of the preceding activity rather than complete removal of the gap, is created. Yet, paradigm~\#3 is applied for all gaps.}

\begin{figure}
  \centering
 \includegraphics[width=0.8\textwidth]{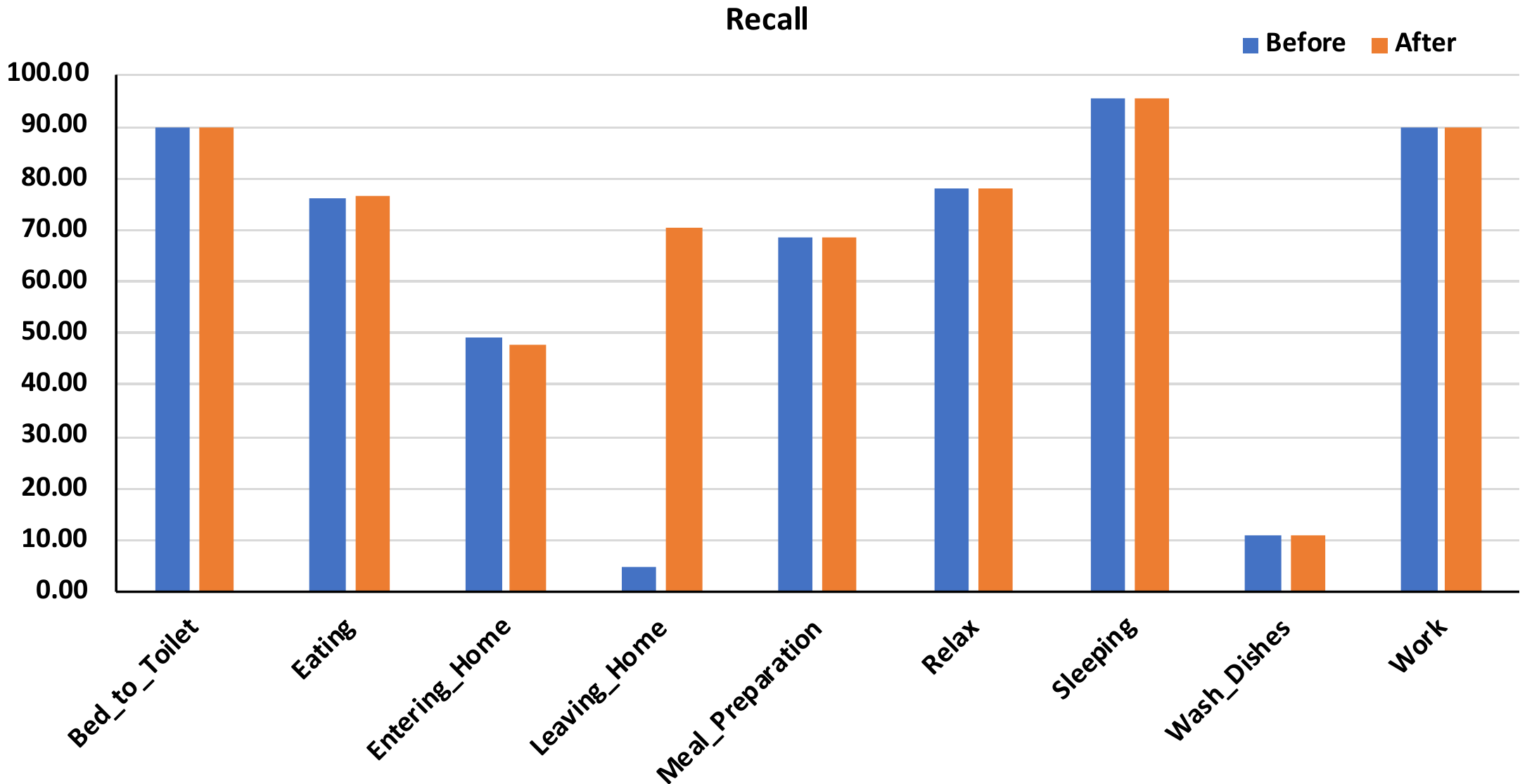}
\caption{Comparison of recall values of paradigm~\#3 before and after applying semantic preprocessing. Note the big improvement that is achieved with the preprocessed activity \textit{Leaving\_Home}.}
\label{fig:RecallN}
\end{figure}

\begin{figure}
  \centering
 \includegraphics[width=0.8\textwidth]{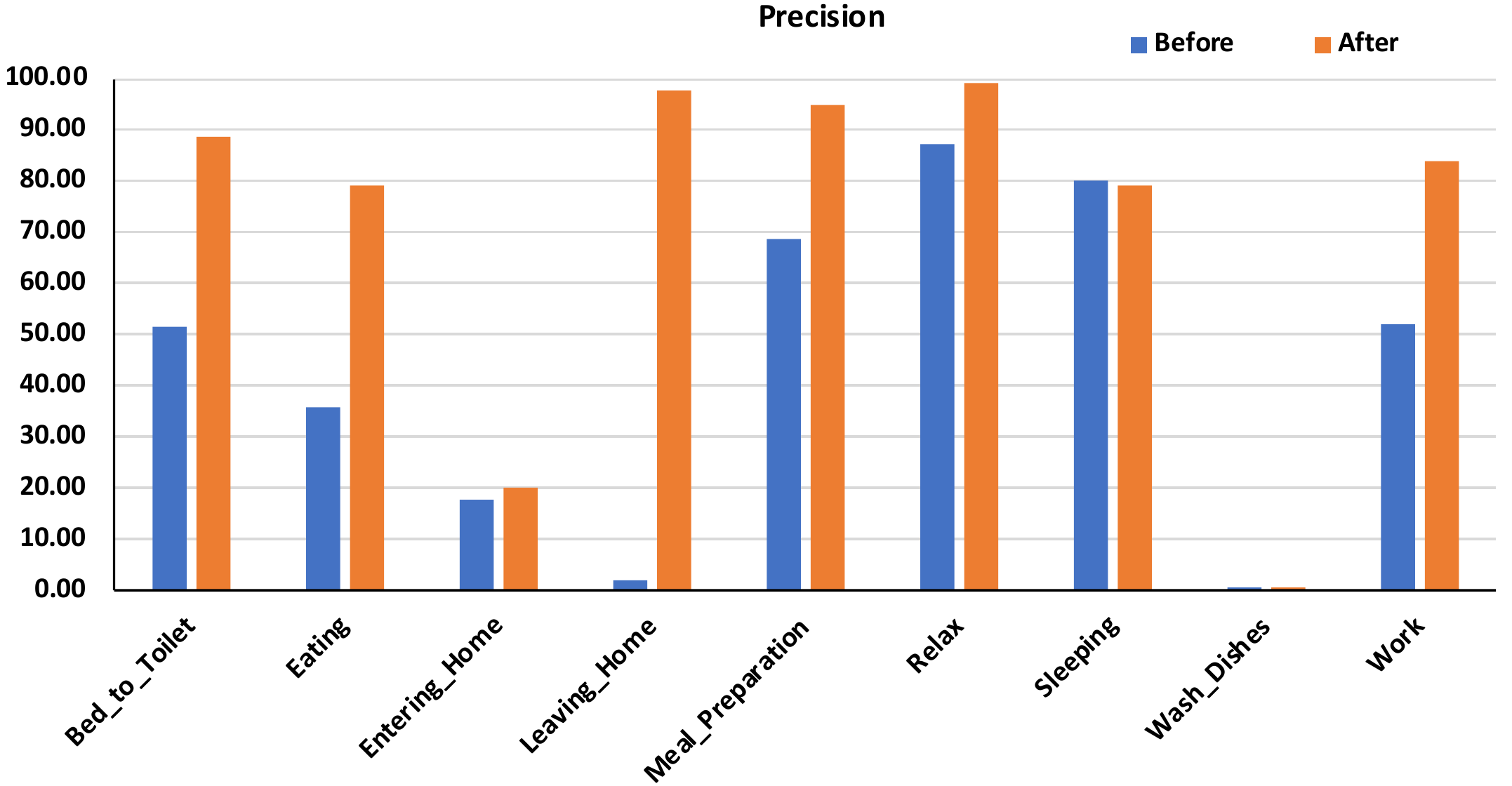}
\caption{Comparison of precision values of paradigm~\#3 before and after applying semantic preprocessing. Burst improvements are achieved with almost all activities, since a considerable source of confusion has been got rid of.}
\label{fig:PrecisionN}
\end{figure}

\begin{figure}
  \centering
 \includegraphics[width=0.8\textwidth]{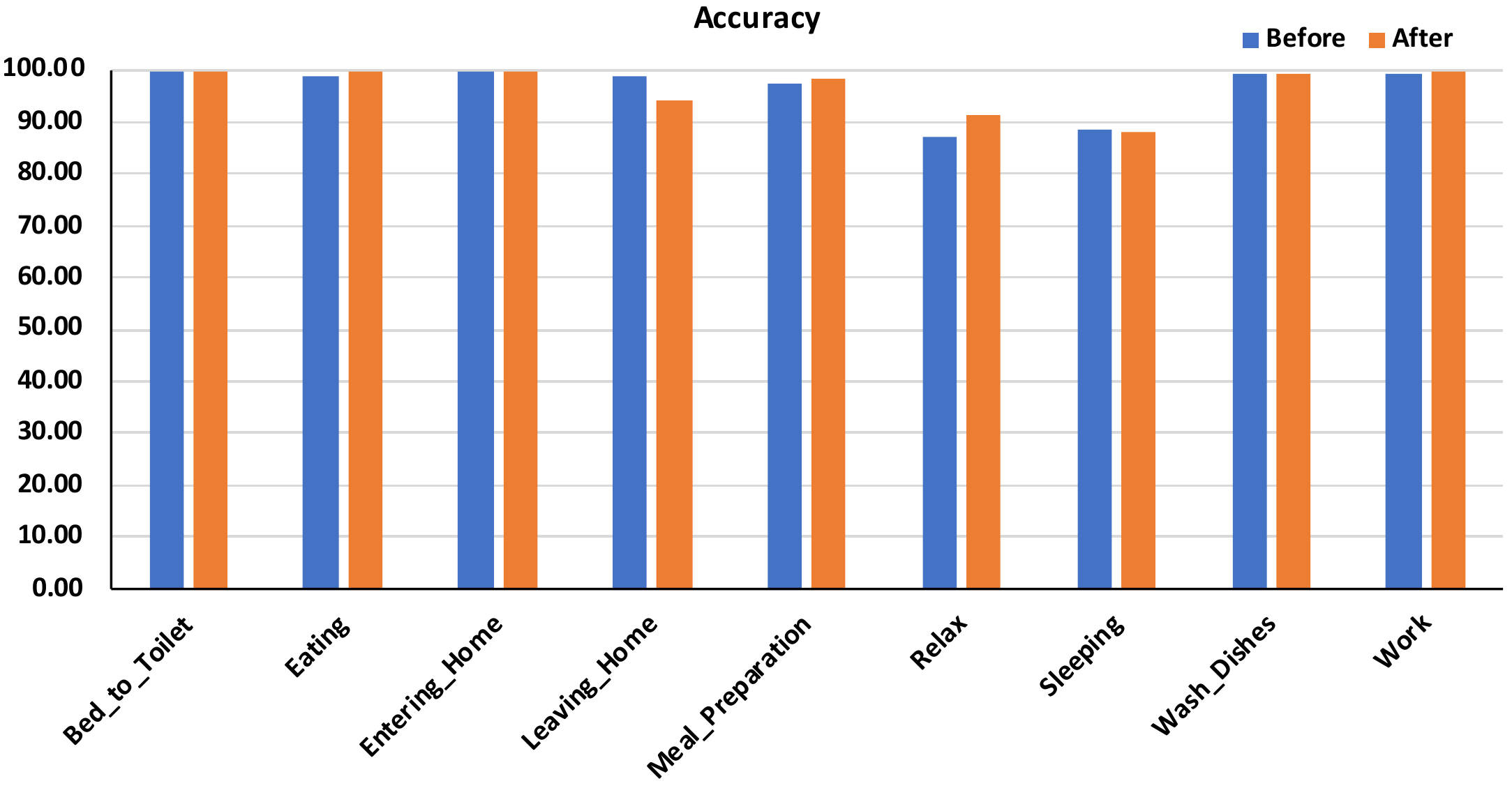}
\caption{\textls[-20]{Comparison of accuracy values of paradigm~\#3 before and after applying semantic~preprocessing.}}
\label{fig:AccuracyN}
\end{figure}

\begin{figure}
  \centering
 \includegraphics[width=0.8\textwidth]{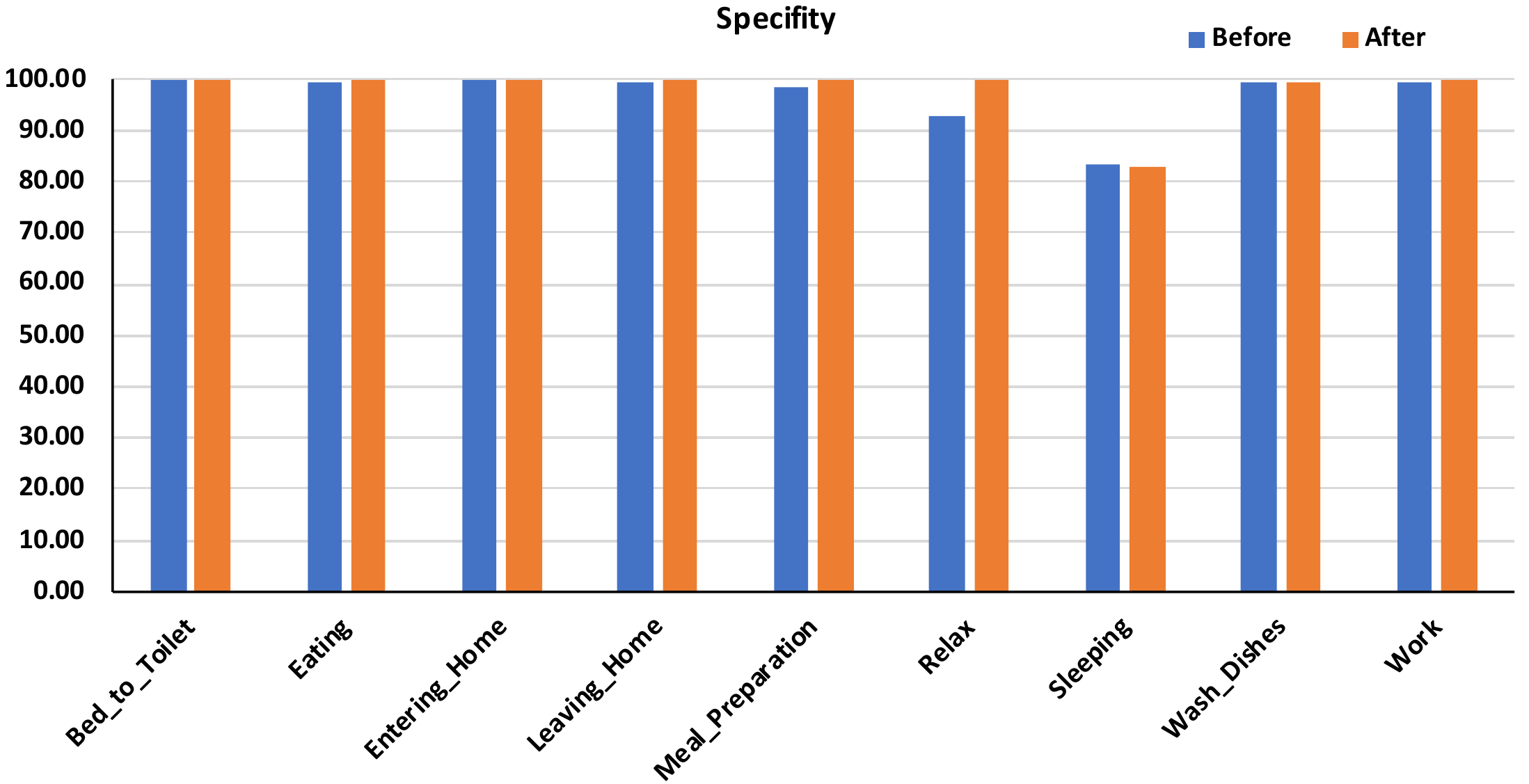}
\caption{Comparison of specificity values of paradigm~\#3 before and after applying semantic~preprocessing.}
\label{fig:SpecificityN}
\end{figure}

To investigate the effect of this hybrid paradigm, we use the aforementioned four metrics for comparison with paradigm~\#3 as the best of the two ingredients of this hybrid paradigm. Figures~\ref{fig:RecallN}--\ref{fig:SpecificityN} show comparison between paradigm~\#3 and the hybrid paradigm. The direct improvement effect appears in the recall of the \textit{Leaving\_Home} activity, Figure~\ref{fig:RecallN}. This is natural because it is the preprocessed activity. The improvement burst is achieved with the precision values, Figure~\ref{fig:PrecisionN}. The reason behind this large improvement for almost all activities is that the presence of obviously redundant annotation gap between \textit{Leaving\_Home} and \textit{Entering\_Home} causing unnecessary confusion to the model estimation process for almost all other activities. When this gap has been removed, the~model became ``more certain'' about the other activities. The remaining two metrics exhibit slight improvement in one of the idle-tended activities, \textit{Relax}, as a result of getting rid of this confusing gap.

From this discussion, we can defend the success of paradigm~\#3 in improving the performance of semi-supervised statistical and learning-based activity recognition systems. If there is a room for annotation gap semantic preprocessing to deal with obvious scenarios, it will be a good addition that pushes performance improvement.

\unskip
\unskip
\unskip

 
\section{Conclusions}\label{sec:conclusion}
We addressed the problem of annotation gap existence in input sensor readings of ADL systems. We investigated three paradigms for handling these gaps. The first paradigm drops the non-annotated gaps from the training data. The second paradigm gives all the instances belonging to these gaps a unique \textit{unknown} or  \textit{do-nothing} label. Inspired by some repeated patterns in these gaps, the third paradigm attributes every gap by its preceding and successor activities. A hybrid combination of a modified version of the first and the third paradigms was proposed by semantic preprocessing of some annotation gaps. The major comparison criterion between the proposed three paradigms is their impact on the overall activity recognition performance of the adopting supervised recognition model. An HMM was used as the adopted learning model for evaluation purposes. Evaluation results showed the superiority of the third paradigm over the other two paradigms in most cases. Furthermore, a~noticeable performance improvement is achieved, if proper semantic preprocessing is conducted. The failure cases of the third paradigm were with gaps whose similar successor activities. The failure cases can be exploited to guide optimal sensor assignment and allocation processes.  

\section*{Acknowledgments} 
The authors would like to thank the \textendash~King Abdulaziz City for Science and Technology \textendash~the Kingdom of Saudi Arabia, (Grant number: 13-ENE2356-10).


\bibliographystyle{unsrt}

\end{document}